\documentclass{article}
\pdfoutput=1
\usepackage[preprint]{spconf}
\usepackage{amsmath,graphicx}
\usepackage{amsfonts}
\usepackage[ruled]{algorithm2e}
\usepackage{algorithmic,algorithm2e}
\usepackage{booktabs}
\usepackage{multirow}
\usepackage{color}
\usepackage{floatrow}
\usepackage{comment}
\usepackage{marvosym}
\usepackage{caption}
\usepackage{xcolor}

\newcommand{\highlightChange}{\color{black}}
\def\HC{\highlightChange}

\title{Learning Hypergraphs From Signals With Dual Smoothness Prior}
\name{Bohan Tang\textsuperscript{1}, Siheng Chen\textsuperscript{2,3}, Xiaowen Dong\textsuperscript{1}}
\address{\textsuperscript{1}University of Oxford, \textsuperscript{2}Shanghai Jiao Tong University, \textsuperscript{3}Shanghai AI Laboratory}
\ninept 

\newcommand{\R}{\ensuremath{\mathbb{R}}}

\def\X{\mathbf{X}}
\def\x{\mathbf{x}}
\def\W{\mathbf{W}}

\def\hH{\mathbf{H}}

\def\V{\mathcal{V}}
\def\E{\mathcal{E}}
\def\hhH{\mathcal{H}}

\def\N{\mathcal{N}}
\def\A{\mathbf{A}}

\DeclareMathOperator{\maxx}{max}

\begin{document}
%
\maketitle
%
\begin{abstract}
{\HC Hypergraph structure learning, which aims to learn the hypergraph structures from the observed signals to capture the intrinsic high-order relationships among the entities, becomes crucial when a hypergraph topology is not readily available in the datasets.} There are two challenges that lie at the heart of this problem: 1) how to handle the huge search space of potential hyperedges, and 2) how to define meaningful criteria to measure the relationship between the signals observed on nodes and the hypergraph structure. In this paper, for the first challenge, we adopt the assumption that the ideal hypergraph structure can be derived from a learnable graph structure that captures the pairwise relations within signals. Further, we propose a hypergraph structure learning framework \textbf{HGSL} with a novel \emph{dual smoothness prior} that reveals a mapping between the observed node signals and the hypergraph structure, whereby each hyperedge corresponds to a subgraph with both \textit{node signal smoothness} and \textit{edge signal smoothness} in the learnable graph structure. Finally, we conduct extensive experiments to evaluate \textbf{HGSL} on both synthetic and real world datasets. Experiments show that \textbf{HGSL} can efficiently infer meaningful hypergraph topologies from observed signals.
\end{abstract}
\section{INTRODUCTION}
\label{sec:intro}
High-order relationships involving multiple entities exist in broad domains: multi-particle events in physics~\cite{NEURIPS2020_fb4ab556}, multi-component reactions in chemistry~\cite{Yadati2020NHPNH}, co-authorships in social science~\cite{chien2022you}, etc. Hypergraphs, as the generalization of graphs with pairwise edge relations, are a powerful tool for modelling higher-order relationships among entities. They consist of nodes and hyperedges that can connect an arbitrary number of nodes, where nodes represent entities and hyperedges represent higher-order relationships among nodes.

Learning with hypergraphs has seen an increasing amount of interest in the fields of both machine learning and signal processing~\cite{Yadati2020NHPNH,chien2022you,8887197,xu2022dynamic}. Most of the existing methods assume the hypergraph structure as known prior knowledge. However, a suitable hypergraph is not always readily available in real world applications. Therefore, it becomes important to develop methods to learn the hypergraph structure from signals observed on the nodes to capture the intrinsic high-order relationships, which motivates this work.


Though a hypergraph is a generalization of a graph with pairwise relations, existing graph structure learning methods~\cite{pmlr-v51-kalofolias16,7178669,pu2021learning} cannot be directly used to learn hypergraph structures. Indeed, these methods are based on the node signal smoothness assumption~\cite{8700665,8700659}, which can only explain the formation of pairwise relationships and not that of high-order relationships containing more than two nodes. Recently, a few works~\cite{NEURIPS2020_fb4ab556, xu2022GroupNet} have attempted to learn hypergraphs from node signals, but none of them has fully addressed the two most critical challenges in hypergraph structure learning: 1) how to deal with the huge search space of potential hyperedges, and 2) how to define meaningful criteria to evaluate the relationship between the node signals and the hypergraph topology. The first challenge is from the hypergraph structure property: there are $2^{N}$ potential hyperedges for a hypergraph with $N$ nodes, and such a search space may lead to intractable computational and memory complexity even in the case of hypergraphs with tens of nodes. The second challenge is vital for the design and interpretability of a hypergraph structure learning framework.

In this paper, to address the first challenge, we avoid enumerating all the potential hyperedges by assuming that the ideal hypergraph structure can be derived from a learned graph structure of pairwise relations. Then, for the second challenge, we propose a framework to learn the hypergraph from the node signals under a novel dual smoothness prior that discloses the relationship between the node signals and the hypergraph structure, i.e., each hyperedge corresponds to a subgraph with both node signal smoothness and edge signal smoothness in the learned graph structure. The dual smoothness prior enables the framework to learn hypergraphs where hyperedges are formed by nodes with similar node signals and similar pairwise relations. Finally, we conduct experiments on both synthetic and real world datasets. Results show that our proposed hypergraph structure learning framework outperforms all of the baselines, which confirms the efficacy of the proposed framework.


\section{RELATED WORK}
\label{subsec:rw}
Hypergraphs have been extensively used in analysing node signals with high-order relationships~\cite{Yadati2020NHPNH, chien2022you, 8887197,xu2022dynamic}. However, the problem of learning a hypergraph structure from observed node signals is under-explored. To our knowledge, there exist only two papers~\cite{NEURIPS2020_fb4ab556, xu2022GroupNet} that try to address this problem. \cite{NEURIPS2020_fb4ab556} formulates a $M$-uniform hypergraph with $N$ nodes as a $N^M$-dimensional tensor and employs a neural network to learn this tensor with supervision. This method does not provide any explicit criteria for the relationship between the node signals and the hypergraph topology. Moreover, it also suffers from huge computational and memory complexity. \cite{xu2022GroupNet} learns a hypergraph from node signals for a downstream task by assuming that each node contributes to at least one hyperedge whose internal nodes are highly correlated. This paper defines a criterion to model the relationship between the node signals and the hypergraph structure. However, this criterion requires the learning framework to traverse the hyperedge search space multiple times, which makes it suffer from high computational and memory complexity as well.

In addition to the papers mentioned above, three research topics in the literature are relevant to the hypergraph structure learning problem: hypergraph reconstruction from given network structures~\cite{Young2021HypergraphRF,9022661}, hypergraph structure optimization~\cite{ijcai2022p267,9747687,9001176}, and simplex learning~\cite{9044758}. First, \cite{Young2021HypergraphRF,9022661} study the problem of reconstructing a hypergraph from a given network structure. These two papers, however, study the relationship between pairwise and high-order structures, rather than the relationship between the node signals and the high-order structure. Second, the hypergraph structure optimizationn~\cite{ijcai2022p267,9747687,9001176} aims at learning an optimal hypergraph structure for the downstream tasks based on the given node signals and hypergraph structure. This problem differs from hypergraph structure learning in two perspectives: 1) it starts with the given node signals and hypergraph structure, whereas in hypergraph structure learning our input is merely the observed node signals; and 2) it only cares about the performance of downstream tasks and does not necessarily the quality of the learned hypergraph structure, whereas in hypergraph structure learning we have no downstream tasks and are only interested in the learned structure. Finally, the goal of simplex learning~\cite{9044758} is to learn simplicial complexes from the given node signals. The simplicial complex is the simplest possible polytope made with line segments in any given dimension{\HC, which is closed under taking subsets.} Compared with a simplicial complex, the structure of a hyperedge is much more flexible. In real world applications, a hyperedge can be any set of nodes interacting for a common objective, such as a group of molecules interacting together for a particular reaction to occur. Hence, the learning criteria used in simplex learning cannot be applied to hypergraph structure learning.


\section{Methodology}
\label{sec:method}
\subsection{Problem Formulation}
\label{subsec:pf}
Let $\hhH = \{\V, \E, \hH \}$ be a hypergraph, where $\V$ is the node set with $|\V| = N$, $\E$ is the edge set and $\hH \in \{0, 1\}^{|\V|\times |\E|}$ is the incidence matrix embedding the hypergraph structure where $\hH_{ij} = 1$ indicates that hyperedge $j$ contains node $i$. The observed node signals are denoted as $\X =[\x_1, \x_2, ..., \x_N]^{\top} \in \R^{|\V| \times d_f}$, where $\x_i$ is the signals of node $i$ and $d_f$ is the number of signals observed. In this paper, we assume that there are no duplicate hyperedges in $\hhH$. Our goal is to design a framework to learn a hypergraph structure $\hH$ that captures the high-order relationships of nodes reflected in the signals $\X$ without traversing the entire hyperedge search space.

\subsection{Learning Assumption}
\label{subsec:lc}
In this section, we present our learning assumption that the hypergraph structure can be derived from a learned graph of pairwise relations.
To make our discussion accessible, we present a probabilistic analysis framework, in which the relationship between the node signals $\X$ and the hypergraph structure $\hH$ is represented by a distribution $p(\hH|\X)$ and the hypergraph structure learning problem is formulated as a maximum a posteriori (MAP) estimation problem:
\begin{equation}
\setlength{\abovedisplayskip}{4pt}
\setlength{\belowdisplayskip}{4pt}
\hH^{\star} = \mathop{\maxx}\limits_{\hH}~ p(\hH|\X),
\label{eq:map1}
\end{equation}
where $\hH^{\star}$ is the target hypergraph structure. Assuming that the formation of hyperedges are independent from each other, the MAP problem in Eq. (\ref{eq:map1}) 
can be rewritten as:
\begin{equation*}
\setlength{\abovedisplayskip}{4pt}
\setlength{\belowdisplayskip}{4pt}
\hH^{\star} = \mathop{\maxx}\limits_{\hH}~ \mathop{\Pi}\limits_{j=1}^{2^N} p(\hH_j|\X),
\end{equation*}
where $p(\hH_j|\X)$ is the probability of existence of hyperedge $\hH_j$ conditioned on $\X$. However, this idea suffers from the high-dimensionality brought by the structural characteristics of the hypergraph, namely the fact that given $N$ nodes there are $2^N$ potential hyperedges. This suggests that even hypergraphs with only tens of nodes have billions or trillions of potential hyperedges, and such a large number of potential hyperedges would lead to intractable computational and memory complexity.

In order to design a more efficient hypergraph structure learning framework, we assume that each hyperedge in the hypergraph consists of nodes with pairwise relations. {\HC The intuition behind this assumption is that, in real world applications, a hyperedge is usually a set of nodes interacting in a pair-wise manner and together towards a common goal, e.g., authorship of scientific papers.} On the basis of this assumption, we first model pairwise relations among all nodes by a graph structure $\W\in\R_{+}^{N\times N}$, where $\W_{ij}$ captures the pairwise relationship between node $i$ and node $j$. Furthermore, we utilise a distribution $p(\W|\X)$ to model the relationship between $\X$ and the graph structure $\W$. After introducing the graph structure, we have:
\begin{eqnarray}
\setlength{\abovedisplayskip}{4pt}
\setlength{\belowdisplayskip}{4pt}
\hH^\star & = & \mathop{\maxx}\limits_{\hH}~\int p(\hH, \X, \W)d\W  \nonumber\\
& = & \mathop{\maxx}\limits_{\hH}~\int p(\hH|\X, \W)p(\X|\W)p(\W) d\W \nonumber \\
& = & \mathop{\maxx}\limits_{\hH}~\int p(\hH|\X, \W)p(\W|\X)d\W .
\label{eq:inter1}
\end{eqnarray}

\begin{figure}[t] 
\centering
\includegraphics[width=0.75\textwidth]{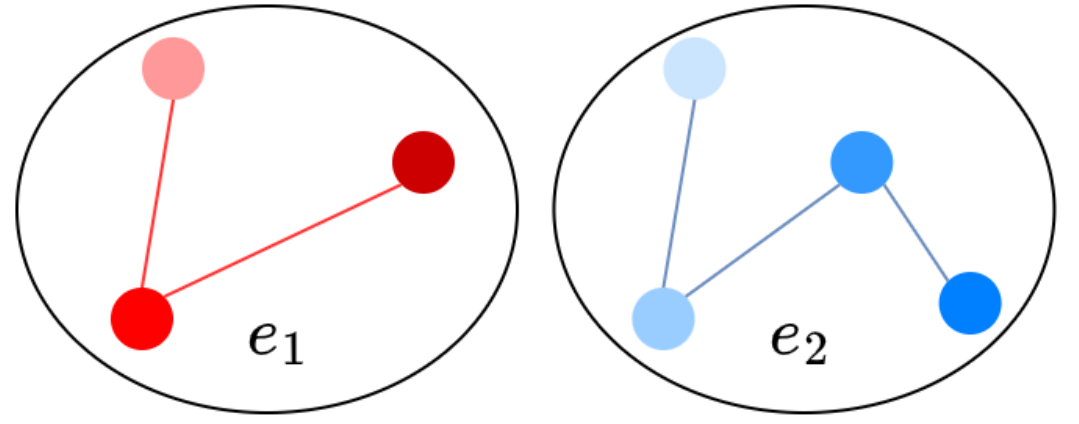}
\vspace{-1mm}
\caption{A hypergraph that obeys the dual smoothness prior. $e_1$ and $e_2$ are two different hyperedges. Nodes of similar colour have similar node signals, and edges of similar colour have similar edge signals.}
\label{fig:hyperedge_exmaple}
\end{figure}

For high-dimensional matrices $\W$, the computation of the integral on the right-hand side of Eq.~(\ref{eq:inter1}) is extremely hard. To simplify the problem, inspired by~\cite{Olshausen1997SparseCW}, we impose that when $\X$ is given we only care about the learnable structure $\W^{\star}$ that maximises $p(\W|\X)$. Therefore, Eq.~(\ref{eq:inter1}) can be simplified as:
\begin{eqnarray}
\label{eq:inter2}
\setlength{\abovedisplayskip}{4pt}
\setlength{\belowdisplayskip}{4pt}
\hH^\star &=& \mathop{\maxx}\limits_{\hH} p(\hH|\X, \W^\star)p(\W^\star|\X), \\
\text{s.t.} &~& \W^\star = \mathop{\maxx}\limits_{\W}~p(\W|\X).\nonumber
\end{eqnarray}
Since $p(\W^\star|\X)$ is a constant when $\W^\star$ and $\X$ are given, Eq. (\ref{eq:inter2}) can be further simplified as:
\begin{eqnarray}
\setlength{\abovedisplayskip}{4pt}
\setlength{\belowdisplayskip}{4pt}
\hH^\star &=& \mathop{\maxx}\limits_{\hH} p(\hH|\X, \W^\star),
\label{eq:final} \\
\text{s.t.} &~& \W^\star = \mathop{\maxx}\limits_{\W}~p(\W|\X).\nonumber
\end{eqnarray}
Accordingly, solving the hypergraph structure learning problem is equal to solving two sub-problems: 1) learn a graph structure $\W^{\star}$ from $\X$, and 2) obtain hyperedges in $\hH^{\star}$ based on $\W^{\star}$ and $\X$.

The reformulation of Eq. (\ref{eq:map1}) introduced above allows us to design a framework to learn the hypergraph structure without enumerating all the potential hyperedges. The hypergraph structure learning framework is elaborated in the next section.

\subsection{Learning Framework}
\label{subsec:lf}
In this section, we first introduce the dual smoothness prior, which is the basis of our learning framework. Then, under the dual smoothness prior, we design a learning framework to solve the sub-problems shown in Section~\ref{subsec:lc}. The framework has two steps: a graph structure learning step, and a line graph community detection step.

\textbf{Dual smoothness prior.} To solve the sub-problems discussed in Section~\ref{subsec:lc}, we propose a dual smoothness prior to measure the relationship between the node signals and the hypergraph structure, namely, a hyperedge corresponds to a subgraph with both node signal and edge signal smoothness in the learnable graph structure. Here, node signal and edge signal smoothness mean that spatially close-by nodes/edges should have similar signals. 
As an example, consider a co-authorship hypergraph where nodes are researchers, and a hyperedge is a paper. In this case, node and edge signals can represent the features of individual researchers and the strength of ties among them, respectively. It is reasonable to assume that related researchers have similar features (node signal smoothness), and {\HC the common academic interests shared by the researchers lead to similar strength of pairwise ties (edge signal smoothness).} We set the edge signals as the edge weights in the learnable graph structure $\W^\star$.

\textbf{Graph structure learning (GSL) step.} The GSL step is designed to address the first sub-problem based on the node signal smoothness of the proposed dual smoothness prior. Inspired by previous graph structure learning methods~\cite{pmlr-v51-kalofolias16,7178669,pu2021learning}, we assume the strength of the relationship between two nodes is proportional to the similarity between these nodes, which can be formulated as the following objective function:
\begin{equation}
\setlength{\abovedisplayskip}{4pt}
\setlength{\belowdisplayskip}{4pt}
    \sum_{i,j}\W_{ij}||\x_i - \x_j||^2_2 - \alpha \mathbf{1}\log(\W\mathbf{1}^{\top}) + \beta||\W||^2_2,
\label{eq:gl}
\end{equation} 
where $\mathbf{1} =[1, 1, ..., 1 ]\in\R^{1\times N}$ is an all-one row vector, $\alpha$ and $\beta$ are two hyperparameters. In Eq. (\ref{eq:gl}), the first term reflects the smoothness of the node signals on the graph, while the second and third terms promote the connectivity and uniform edge weights of the graph structure respectively. In practice, we use the primal-dual splitting (PDS) algorithm~\cite{pmlr-v51-kalofolias16} to learn $\W^{\star}$ by minimizing Eq. (\ref{eq:gl}).

\begin{figure}[t] 
\centering
\includegraphics[width=0.7\textwidth]{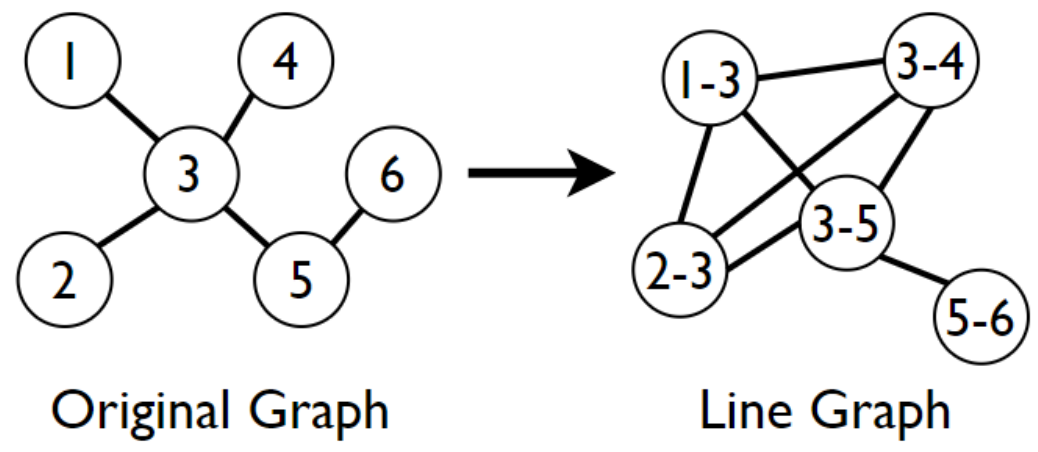}
\vspace{-2mm}
\caption{An illustration of the line graph construction procedure.}
\label{fig:line_graph_example}
\end{figure}


\textbf{Line graph community detection (LGCD) step.} The LGCD step aims to address the second sub-problem. The node signal smoothness of the dual smoothness prior is captured by the connectivity of the graph learned in the GSL step, where only nodes with smooth signals are connected. Therefore, the second sub-problem can be formulated as setting connected subgraphs in $\W^\star$ with edge signal smoothness (as illustrated in Fig.~\ref{fig:hyperedge_exmaple}) as hyperedges in $\hH^\star$. To capture both the connectivity and the edge signals in the graph structure $\W^{*}$, we construct a line graph~\cite{9431673} corresponding to the graph structure $\W^{*}$, where each node is an edge in $\W^{*}$ and an edge between two nodes in the line graph means the two corresponding edges in $\W^{*}$ share a common end node. Let $\mathcal{L}(\W^{*}) = \{\V^{l}, \A^{l}\}$ be the line graph, where $\V^{l}$ is the node set with $|\V^l|=L$, and $\A^{l}\in\{0,1\}^{L\times L}$ is the adjacency matrix where $\A^{l}_{ij} = 1$ indicate that node $i$ and node $j$ are connected. We also let $\x^{l}\in\R_{+}^{L}$ contain node signals and $\x^{l}_i$ be the signal of node $i$ in $\mathcal{L}(\W^\star)$ that is its corresponding edge signal in $\W^{*}$. Fig.~\ref{fig:line_graph_example} shows an example of the construction of a line graph.

With the line graph, the hyperedges, fitting the dual smoothness prior, are equal to the connected subgraphs with smooth node signals now in $\mathcal{L}(\W^\star)$. To find these subgraphs in $\mathcal{L}(\W^\star)$, we first 
compute a weighted adjacency matrix $\W^l\in\R^{L\times L}_{+}$ for $\mathcal{L}(\W^{*})$:
\begin{equation*}
\setlength{\abovedisplayskip}{4pt}
\setlength{\belowdisplayskip}{4pt}
    \W^l_{ij} = \exp(-\frac{||\x^{l}_i-\x^{l}_j||_{2}^{2}}{2}) \cdot \A^{l}_{ij},
\end{equation*}
where the value of $\exp(-\frac{||\x^{l}_i-\x^{l}_j||_{2}^{2}}{2})$ is proportional to the signal similarity between node $i$ and node $j$ in $\mathcal{L}(\W^\star)$, and $\W^l_{ij}$ is the product of $\exp(-\frac{||\x^{l}_i-\x^{l}_j||_{2}^{2}}{2})$ and $\A^{l}_{ij}$. Based on $\W^l$, the LGCD step detects communities in $\mathcal{L}(\W^{*})$ and sets these communities as hyperedges to form $\hH^{\star}$. To ensure that every detected community is connected, in practice, we employ the Leiden algorithm~\cite{V2019From}, a state-of-the-art community detection algorithm, to detect communities in $\W^l$. The learning framework is termed hypergraph structure learning (HGSL) and is summarized in Algorithm~\ref{al:two_step}. 

\begin{algorithm}[t!]
    \caption{Hypergraph Structure Learning With Dual Smoothness Prior (\textbf{HGSL})}
\label{al:two_step}
     \begin{algorithmic}[1]
     \renewcommand{\algorithmicrequire}{\textbf{Input:}}
     \renewcommand{\algorithmicensure}{\textbf{Output:}}
     \REQUIRE Signal $\X$.
     \ENSURE Hypergraph incidence matrix $\hH^\star$.

    {\STATE  {\bfseries GSL step}: \\     
    $\bullet$ Minimize Eq.~(\ref{eq:gl}) to learn a graph structure $\W^{\star}$ of pairwise relations.
    }

    {\STATE  {\bfseries LGCD step}: \\     
    $\bullet$ Construct the line graph $\mathcal{L}(\W^\star)$ corresponding to the graph structure $\W^{\star}$. Please refer to the main text for the details of the construction of $\mathcal{L}(\W^\star)$.\\
    \vspace{0.1cm}
    $\bullet$ Compute the weighted adjacency matrix $\W^l\in\R_{+}^{L\times L}$ of $\mathcal{L}(\W^\star)$, where $\W^l_{ij} = \exp(-\frac{||\x^{l}_i-\x^{l}_j||_{2}^{2}}{2}) \cdot \A^{l}_{ij}$.\\
    \vspace{0.1cm}
    $\bullet$ Use the Leiden algorithm to find communities based on $\W^l$, and set these communities as hyperedges in $\hH^{\star}$.}
    \end{algorithmic}     
\end{algorithm}

\textbf{Remark.} We remark on \textbf{HGSL} from three perspectives.
\textit{[Model complexity:]} {\HC Let $M$ be the number of edges in the graph learned by the GSL step. The complexity of the GSL step is $\mathcal{O}(N^2)$ and the complexity of the LGCD step is $\mathcal{O}(M^2)$. Thus, the complexity of \textbf{HGSL} is $\mathcal{O}(N^2 + M^2)$.} 
\textit{[Overlapping rate of a hypergraph:]} We introduce the overlapping rate, an important factor for the performance of \textbf{HGSL}. 
The overlapping rate of a hyperedge is defined as the ratio of nodes in the hyperedge that are involved in more than one hyperedge to the total number of nodes in the hyperedge. Then, the overlapping rate of a hypergraph is the average overlapping rate of its hyperedges. The overlapping rate of a hypergraph impacts \textbf{HGSL} in two ways: 1) a hypergraph with a large overlapping rate might have complex pairwise relations between nodes, which would make the graph structure learning problem solved in the GSL step difficult; and 2) a large overlapping rate may make the signals and pairwise relations of nodes in different hyperedges similar, which might increase the difficulty of the LGCD step. 
\textit{[Difference from prior work:]} {\HC Compared to the community/clique detection models, \textbf{HGSL} only takes the observed node signals as input, rather than the graph structure.} Moreover, the LGCD step of \textbf{HGSL} is conceptually similar to the works on hypergraph reconstruction~\cite{Young2021HypergraphRF,9022661}. However, the LGCD step learns the hypergraph from the node signals and the learnable graph structure $\W^\star$ via our dual smoothness prior instead of setting nodes with dense within connections, i.e., communities and cliques, as hyperedges. Note that we test the impact of the overlapping rate to \textbf{HGSL} and compare \textbf{HGSL} with the framework using community/clique detection in experiments.

\section{Experiments}
\subsection{Experimental settings}
We test our proposed hypergraph structure learning method \textbf{HGSL} by comparing it with four baselines on three metrics, \textit{Recall}, \textit{Precision} and \textit{F1 Score}, in both synthetic and real world datasets.

\textbf{Synthetic dataset.} To test the proposed method in datasets with different scales, we conduct experiments on two synthetic datasets, one containing 20 nodes per hypergraph and the other containing 200 nodes per hypergraph. Each dataset involves 96 different hypergraphs with 5000 signals per node, divided into three parts: 32 hypergraphs with 0\% overlapping rate, 32 hypergraphs with 10\% overlapping rate, and 32 hypergraphs with 25\% overlapping rate. The generation of each synthetic hypergraph contains two steps: structure generation and signal generation. The structure generation algorithm is based on the method proposed in~\cite{do2020structural}. This algorithm iteratively generates new hyperedges until it can no longer generate hyperedges with an overlapping rate less than a preset maximum overlapping rate. The signal generation makes hyperedges with smooth node signals and smooth edge weights. It generates a connected subgraph with smooth edge weights in each hyperedge and integrates the edge weights in every hyperedge to form a precision matrix $\mathbf{\Gamma}$ of a Gaussian distribution. Then, node signals $\X$ are sampled from the Gaussian distribution $\N\!(\mathbf{0},\mathbf{\Gamma})$. 



\textbf{Real world dataset.} We conduct experiments on a real world co-authorship hypergraph with 508 nodes and 269 hyperedges, where a node is an author, a hyperedge is a paper published in an ACM conference, and each node has 1830 signals that are provided by a bag-of-words model with 1830 chosen keywords. This real world hypergraph dataset is based on a dataset preprocessed by~\cite{han2019}.

\textbf{Baselines.} We choose GroupNet~\cite{xu2022GroupNet}, clustering~\cite{8425790}, community detection~\cite{10.1145/3269206.3271697}, and clique detection~\cite{Cazals2008ANO} as our baselines. GroupNet~\cite{xu2022GroupNet} and clustering~\cite{8425790} are methods that use different hypergraph structure learning criteria than \textbf{HGSL}. Clustering~\cite{8425790} makes each hyperedge consist of nodes with smooth signals. GroupNet~\cite{xu2022GroupNet} assumes that each node contributes to at least one hyperedge whose internal nodes are highly correlated. Community detection~\cite{10.1145/3269206.3271697} and clique detection~\cite{Cazals2008ANO} cannot solve the hypergraph structure learning problem directly because they need to start with a graph structure. For comparison, we utilize them only for the second step of our proposed method. Note that, we only run GroupNet in the synthetic dataset with 20 nodes, because, for $N$ given nodes, the computational complexity of GroupNet is around $\mathcal{O}(2^N)$.


\subsection{Results}
\begin{figure}[t] 
\centering
\includegraphics[width=.9\textwidth]{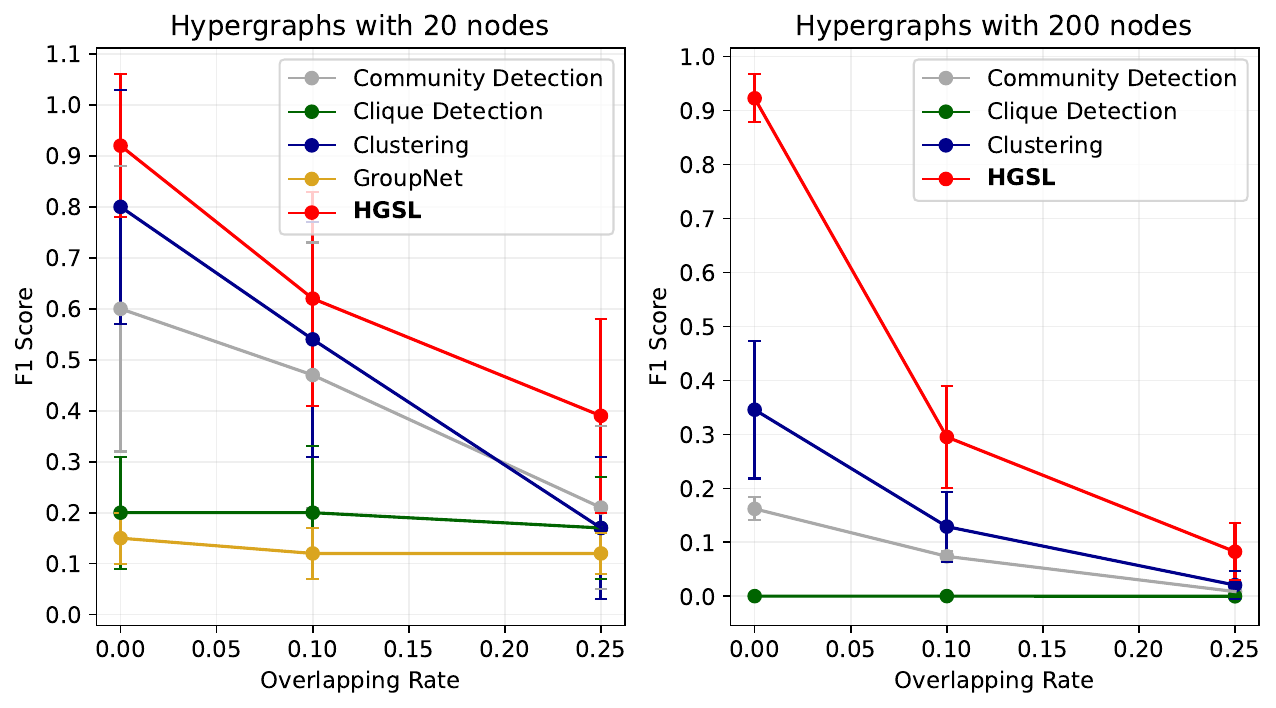}
\vskip -0.05in
\caption{Performance of the proposed method and baselines with respect to the overlapping rate.
}
\label{fig:synthetic_results}
\vskip -0.1in
\end{figure}

\textbf{Synthetic dataset.} We first compare \textbf{HGSL} with the chosen baselines on the synthetic datasets. The quantitative results on the dataset with 20 nodes and 0\% overlapping rate are shown in Table~\ref{tab:syn_20_0}, and the F1 Score on all of the synthetic datasets are summarized in Fig.~\ref{fig:synthetic_results}. Here, we show the average and the standard deviation of the chosen metrics computed over 32 hypergraphs of each synthetic dataset. These results lead to two observations. First, \textbf{HGSL} outperform all the chosen baselines in terms of the chosen metrics. This demonstrates that \textbf{HGSL} can learn hypergraphs obeying our proposed criterion more accurately than baselines, which confirms its effectiveness. Second, the F1 Score of all the methods drops as the overlapping rate of the hypergraph increases. This suggests that the overlapping rate of the target hypergraph may be a key factor influencing the performance of the hypergraph structure learning method.

\textbf{Real world dataset.} The results on the real world hypergraph dataset are provided in Table~\ref{tab:real_world}. These results show that \textbf{HGSL} outperform all the baselines in terms of all the chosen metrics, which indicates that our proposed hypergraph structure learning criterion is more effective in this real world hypergraph than the criteria used in the baselines. The learning criteria of the baselines require signals of every pair of nodes in a hyperedge to be smooth. However, in this dataset, a paper can be written by researchers from different areas, where some pairs of nodes might not have smooth signals. On the basis of the proposed dual smoothness prior, \textbf{HGSL} allows some pairs of nodes within a hyperedge to have different signals, which enables it to find the hyperedges not densely connected in the graph learned by the GSL step, e.g., a path in the learned graph.

\begin{table}[h]
\begin{center}
\caption{Results on the synthetic dataset with 20 nodes and 0\% overlapping rate.
}
\label{tab:syn_20_0}\resizebox{\columnwidth}{!}{
\begin{tabular}{c|ccc}
\hline
Method & Recall & Precision & F1 Score
\\
\hline
Community Detection \cite{10.1145/3269206.3271697} & 0.61$\pm$0.23 
&0.61$\pm$0.23 &0.61$\pm$0.23
\\
Clique Detection \cite{Cazals2008ANO} & 0.33$\pm$0.17
&0.15$\pm$0.08 &0.20$\pm$0.11
\\
Clustering \cite{8425790} &0.80$\pm$0.28 &0.80$\pm$0.28 &0.80$\pm$0.28
\\
GroupNet \cite{xu2022GroupNet} &0.80$\pm$0.20 &0.08$\pm$0.03 &0.15$\pm$0.05
\\
\textbf{HGSL} & \textbf{0.95$\pm$0.09}&\textbf{0.90$\pm$0.18}&\textbf{0.92$\pm$0.14}
\\
\hline
\end{tabular}}
\end{center}
\vskip -0.3in
\end{table}

\begin{table}[h]
\caption{Results on real world dataset.
}
\label{tab:ab_in_multi}
\begin{center}\resizebox{\columnwidth}{!}{
\begin{tabular}{c|ccc}
\hline
Method & Recall & Precision & F1 Score
\\
\hline
Community Detection \cite{10.1145/3269206.3271697} & 0.384
&0.384 &0.384 
\\
Clique Detection \cite{Cazals2008ANO} & 0.270
&0.171 &0.209
\\
Clustering \cite{8425790} &0.351 &0.351 &0.351 
\\
\textbf{HGSL} & \textbf{0.454}&\textbf{0.436}&\textbf{0.445}
\\
\hline
\end{tabular}}
\end{center}
\label{tab:real_world}
\vskip -0.3in
\end{table}

\section{CONCLUSIONS}
We propose a hypergraph structure learning framework \textbf{HGSL} to learn hypergraph structures from node signals. \textbf{HGSL} handles the big search space of potential hyperedges by conditioning the ideal hypergraph structure on a learnable graph of pairwise relations. Further, \textbf{HGSL} learns the hypergraph structure based on a novel dual smoothness prior, i.e., each hyperedge corresponds to a subgraph with both node signal and edge signal smoothness in the learnable graph. Our experiments demonstrate the effectiveness of our algorithm in inferring meaningful hypergraph topologies. We believe that \textbf{HGSL} contributes to the fast-growing fields of hypergraph machine learning and hypergraph signal processing where one is interested in analysing signals with high-order relationships whose structure is not explicitly available.



\bibliographystyle{IEEEbib}
\bibliography{refs}

\end{document}